# Find Everything: A General Vision Language Model Approach to Multi-Object Search

Daniel Choi*, Angus Fung*, Haitong Wang*, Aaron Hao Tan*
* Equal contributions

*Abstract*—Efficient navigation and search in unknown environments for multiple objects is a fundamental challenge in robotics, particularly in applications such as warehouse management, domestic assistance, and search-and-rescue. The Multi-Object Search (MOS) problem involves navigating to a sequence of locations to maximize the likelihood of finding target objects while minimizing travel costs. In this paper, we introduce a novel approach to the MOS problem, called Finder, which leverages vision language models (VLMs) to locate multiple objects across diverse environments. Specifically, our approach introduces multi-channel score maps to track and reason multiple objects simultaneously during navigation, along with a score map technique that combines scene-level and object-level semantic correlations. We validate our approach through extensive experiments in both simulated and real-world environments. The results demonstrate that Finder outperforms existing multi-object search methods using deep reinforcement learning and VLM. Additional ablation and scalability studies highlight the importance of our design choices and show the system's robustness with increasing number of target objects. Website: https://find-all-my-things.github.io/

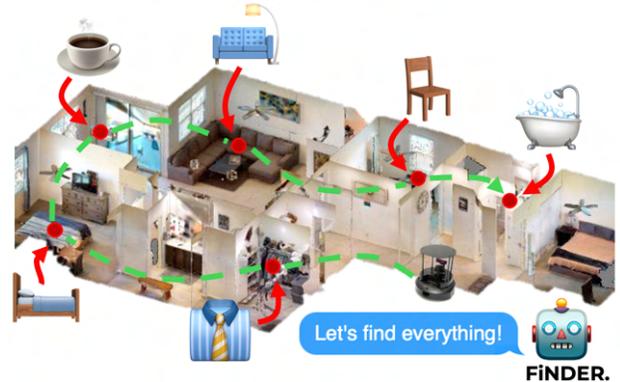

**Fig. 1.** Finder utilizes a vision language model to address the multi-object search problem for a mobile robot in any indoor environment.

## I. INTRODUCTION

In various real-world applications, robots need to efficiently identify and locate multiple objects to complete tasks. This is critical in domains such as warehouse management [1], construction inspection [2], and domestic [3], and retail assistance [4], [5]. This challenge is addressed by multi-object search (MOS) [6] which describes the problem of finding multiple target objects in an unknown environment, while minimizing the robot's travel distance and search time [7].

Existing MOS methods can be categorized into: 1) probabilistic planning (PP) [6], [8]-[10], and 2) deep reinforcement learning (DRL) methods [7], [11]-[17]. PP methods address uncertainty in object locations using Partially Observable Markov Decision Processes (POMDPs) to estimate belief states and plan robot actions under partial observability, while DRL methods train robots to explore environments by optimizing action selection using a reward function [18]. However, these methods are limited by: 1) inefficient exploration due to a lack of direct semantic modeling between target objects and the scene [15], and 2) poor generalizability due to the sim-to-real gap [16].

Recently, large foundation models (LFMs) [19], such as vision-language models (VLMs), and large language models (LLMs), have demonstrated strong common-sense knowledge and reasoning abilities. These models can address the limitations of existing MOS methods by explicitly modeling the semantic correlation between target objects and the environment, while also generalizing across environments. As a result, these models have been applied to single object search (SOS) tasks using either: 1) VLMs (e.g., CLIP, BLIP, etc.) to generate scene-level embeddings that capture the semantic correlations between the robot's environment, and the target object [16], [20]-[23], to guide the robot towards regions with high target object likelihood; or, 2) VLMs/LLMs to generate scene captions that describe both the spatial layout and semantic details of the robot's environment [24]-[30], which are then used to plan the robot's actions. However, these SOS methods have limitations: 1) they cannot be directly applied to MOS, as they lack explicit mechanisms to track and reason about multiple objects simultaneously, and 2) scene-level embeddings are often noisy and coarse [31], which cannot be effectively applied in object-dense environments. In such cases, fine-grained, object-level embeddings are needed.

In this paper, we introduce Finder, the first MOS approach that leverages VLMs to locate multiple target objects in various unknown environments, Fig. 1. Our key contributions are: 1) We introduce a multi-channel score representation that simultaneously captures and tracks the semantic relationships between multiple target objects, the environment, and scene objects. This addresses the limitations of existing MOS methods by explicitly modeling how objects relate to their surroundings to improve exploration efficiency; 2) We develop a novel score map computation technique that combines scene-to-object correlations, which capture how each target relates to the overall scene, with object-to-object correlations that model contextual relationships between detected objects and targets. This overcomes the limitations of coarse scene-level embeddings used by prior VLM-based SOS methods; and 3) We conduct extensive benchmark simulation and real-world experiments to validate Finder's performance.

## II. RELATED WORKS

Current object search methods for mobile robots can be categorized into: 1) PP methods for MOS [6], [8]-[10] 2) DRL methods for MOS [7], [11]-[17], and 3) VLM methods for SOS [20]-[30], [32], [33].

### A. Probabilistic Planning Methods (PP) for MOS

PP methods for MOS account for uncertainty in object locations and robot perception by using probabilistic frameworks to estimate belief states and plan actions under partial observability [6]. These methods generally assumed no prior knowledge of object locations, requiring the robot to iteratively update its belief using noisy sensor data. POMDPs are commonly used to address the uncertainty and partial observability in MOS. Usages of POMDPs included: 1) structuring the belief space based on objects and object classes for belief updates across multiple objects [6], 2) using point clouds to construct a occupancy octree for occlusion-aware searches and continuous belief updates [8], 3) managing dynamic environments through belief tree reuses [9], and 4) reducing computational complexity by segmenting the search areas into regions [10]. Simulated experiments were conducted in 2D grid worlds [6], [10], and 3D indoor environments [8], [9]. Real-world experiments were conducted in indoor environments using robots such as Spot and Kinova MOVO [6], [8].

### B. Deep Reinforcement Learning Methods (DRL) for MOS

In DRL methods for MOS, the robot is trained to explore unknown environments and locate multiple objects by repeatedly interacting with the environment during offline training [7]. These methods used DRL frameworks such as Deep Q Networks (DQN) [11], Proximal Policy Optimization (PPO) [7], [12]-[15], [17], or hybrid approaches that combine classical SLAM with learned policies [16], to optimize the robot's navigation action selection based on RGB-D inputs [7], [12]-[17], LiDAR [16], or graph-based data [11]. The outputs of the DRL policies included: 1) discrete navigation actions (e.g., go straight, turn right, etc.) [7], [11], [13]-[15], [17], 2) continuous navigation actions [12], or 3) navigation waypoints [16]. DRL methods were primarily evaluated in simulation environments using Matterport3D [7], [13]-[16], custom-built environments [11], Gibson [14]-[16] and iGibson [12], [17]. Some methods were validated on physical robots, such as LoCoBot [12], [16] or Toyota HSR [12], [17].

### C. Large Foundation Model Methods (LFM) for SOS

LFM methods for SOS focus on enabling robots to navigate unknown environments by leveraging natural language descriptions and visual inputs [23]. These methods incorporate VLMs and LLMs to guide object search using semantic reasoning and multi-modal robot perception [30]. Namely, these methods utilize RGB [21], [22], [25], [26], [33] or RGB-D [20], [23], [24], [27]-[30], [32] images from egocentric robot perspectives, to detect target objects using open-vocabulary models (e.g., Grounding DINO [34], SAM [35]), followed by planning discrete actions such as moving forward or turning. The models used pre-trained VLMs such as CLIP [20], [21], [30], [33], GLIP [24], [25], Llama-Adapter [26], BLIP [22], [23], [28] as well as LLMs such as, GPT-4 [26], GPT-4V [29] [30], DeBERTa [24], RoBERTa [27] for navigation reasoning and instruction parsing. Experiments were conducted in simulated environments such as Habitat [20], [29], [32] RoboTHOR [20], [24], [25], PASTURE [20], [21], [23], [24] HM3D [21]-[24], [26]-[28], [30] HSSD [28], Gibson [21], [23], [27] ProcTHOR-10k [29], [33], were commonly used to test performance in indoor settings. Experiments with real-world hardware, including LoCoBot [20], [32], iRobot [26], Turtlebot [25], [29], [30], Jackal [27], and Spot [23], further validated the proposed approaches in real-world scenarios.

### D. Summary of Limitations

PP methods face computational inefficiency in scaling to large, complex environments due to the need to maintain and update belief states for multiple objects over extended planning horizons [6]. DRL methods are limited by 1) inefficient exploration, as they optimize navigation based solely on sensory inputs without directly modeling semantic correlations between target objects and the scene [14], and 2) poor generalizability, requiring extensive training data and resources that hinder transferring learned policies from simulation to real-world scenarios [16], [36]

While LFM methods can generalize well in a zero-shot manner, they are limited by: 1) their focus on SOS, making them unable to track multiple objects simultaneously for MOS where objects may be semantically related [23], and 2) reliance on coarse embeddings obtained from LFMs that capture only scene-level correlation between target objects and the environment, missing fine-grained correlations between target objects with objects in the scene [20], [29].

To address these limitations, we propose Finder, the first VLM-based approach that introduces multi-channel maps to address the challenges of tracking multiple objects simultaneously for MOS, and a score map technique to capture and fuse both scene and object-level correlations.

## III. THE MULTI-OBJECT SEARCH PROBLEM FORMULATION

The MOS problem requires a mobile robot to search for a list of target objects in an unknown environment. The robot is equipped with an RGB-D camera and has a state $\mathbf{x}_r(t) \in \mathbb{R}^N$ at time $t$, where $\mathbf{x}_r(t) = (x, y, \phi)$ represents its position and orientation on a 2D surface. The environment consists of $L$ static scene objects $\mathcal{O}_{\text{sne}} = \{o_{s_1}, o_{s_2}, ..., o_{s_L}\}$. The set of $K$ static target objects to be located is denoted by $\mathcal{O}_{\text{tgt}} = \{o_{t_1}, o_{t_2}, ..., o_{t_K}\}$, $\mathcal{O}_{\text{tgt}} \subseteq \mathcal{O}_{\text{sne}}$, where each object $o_{t_j}$ occupies an unknown state $\mathbf{x}_{t_j}$. The objective of the MOS problem is to minimize the cumulative distance travelled $d$ required for the robot to locate all objects in $\mathcal{O}_{\text{tgt}}$ given control inputs $u(t)$:

$$\min_{u(t)} \quad d = \int_0^T \|\dot{\mathbf{x}}_r(t)\| \, dt \,, \text{s.t.} \ \sum_{j=1}^{K} \mathbf{1}(o_{t_j}) = K \quad (1)$$

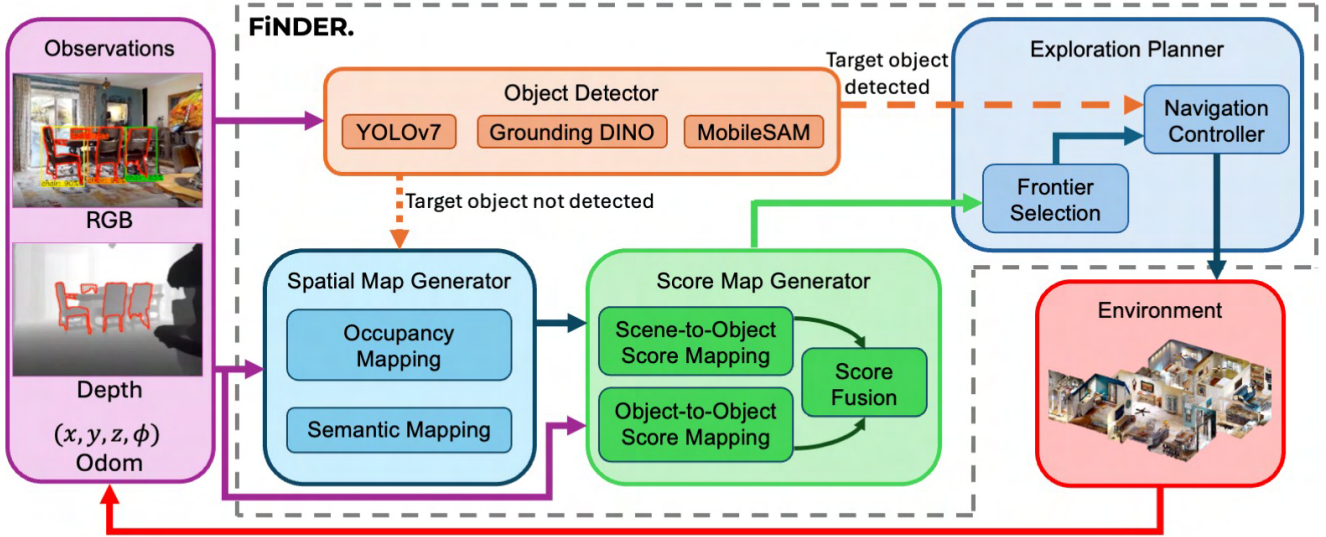

**Fig. 2.** The proposed Finder architecture consists of four modules: 1) *Object Detector* which identifies whether a scene/target object is present in the robot view, 2) *Spatial Map Generator* which generates an occupancy and semantic map for navigation, 3) *Score Map Generator* which generates a unified score map representing the combined scene-to-object score map and object-to-object score map, and 4) *Exploration Planner* which selects the next frontier or target waypoint to navigate towards.

where $T$ is the total time to complete the search, and $\mathbf{1}(\cdot)$ is the indicator function of detecting the target objects.

## IV. THE FINDER ARCHITECTURE

The proposed multi-object search architecture, Finder, is presented in Fig. 2. The goal is for the robot to find multiple target objects in an unknown environment by exploring areas with the highest semantic correlation scores. These scores are based on the scene-level correlation between the target objects and the scene as well as object-level correlation between the scene objects and target objects. This allows the robot to efficiently explore regions that are semantically correlated with the target objects. The architecture consists of four main modules: 1) *Object Detector*, 2) *Spatial Map Generator*, 3) *Score Map Generator*, and 4) *Exploration Planner*.

### A. Object Detector

The *Object Detector* module identifies whether a scene object or a target object is in the robot view from an RGB image $I_{RGB}^t$ and depth image $I_D^t$ at time $t$. Specifically, YOLOv7 [37] is used to output class labels $c_i$ (within the COCO [38] classes) and bounding boxes $b_i$ from $I_{RGB}^t$, [39], while Grounding DINO [34] is used to detect objects not in COCO. We selected YOLOv7 as it incorporates Extended Efficient Layer Aggregation (ELAN) which improves feature learning efficiency, needed for real-time inference for mobile robots. We selected Grounding DINO as it enables open-vocabulary detection, allowing zero shot detection of objects beyond the pre-defined categories of COCO.

Segmentation masks $S^t$ are generated using MobileSAM [35] from the RGB image $I_{RGB}^t$ and bounding boxes $b_i$ from detected scene objects. If a target object $o_{t_j}$ is detected, the closest point on the target relative to the robot is identified from $I_D^t$ and $S^t$. The closest point $p_i$ is then projected into 3D space using the pinhole camera model [40], [41], to obtain a target object waypoint $\mathbf{w}_{t_i}$, which is passed to the *Navigation Controller* within the *Exploration Planner*. If a target object is not detected, the masks $S^t$ are passed into the *Spatial Map Generator* module.

### B. Spatial Map Generator

The *Spatial Map Generator* module generates metric maps of the environment using two sub-modules: 1) the *Occupancy Mapping*, and 2) *Semantic Mapping*. The *Occupancy Mapping* sub-module generates an occupancy map $\mathbf{M}_o^t \in \mathbb{R}^{H \times W}$ from depth image $I_D^t$ and odometry information $\boldsymbol{\rho}^t$ at time $t$, updating as the robot navigates in the environment. Obstacles are identified by converting $I_D^t$ into a point cloud $\mathbf{P}^t$, and projecting these points onto the occupancy map $\mathbf{M}_o^t$. The *Semantic Mapping* module generates a semantic map $\mathbf{M}_s^t \in \mathbb{R}^{L \times H \times W}$ from the RGB and depth images $I_{RGB}^t$ and $I_D^t$, respectively. Specifically, the segmentation mask $S^t$ for each detected scene object is projected into a 2D map using the semantic mapping procedure in [42]. The output maps $\mathbf{M}_o^t$ and $\mathbf{M}_s^t$ are passed into the *Score Map Generator* module.

### C. Score Map Generator

We introduce the *Score Map Generator* module, consisting of two sub-modules: 1) the *Scene-to-Object (StO) Score Mapping*, 2) the *Object-to-Object (OtO) Score Mapping*, and 3) *Score Fusion*. The *StO Score Mapping* generates scene-level correlation scores to capture the semantic relationships between target objects and the scene. The *OtO Score Mapping* generates object-level correlation scores to capture the relationships between target objects and scene objects.

*1) Scene to Object Score Mapping*

The *StO Mapping* module generates a score map where each element represents the semantic correlation of a specific location with respect to each of the target objects, Fig. 3. Specifically, it takes as inputs $I_{RGB}^t$ and $\mathbf{M}_o^t$, and outputs a multi-channel *StO* score map $\mathbf{V}_{s \to o_{tgt}}^t \in \mathbb{R}^{K \times H \times W}$ of the same spatial dimension as the occupancy map. The semantic correlation $S(I_{RGB}^t, o_{t_j})$ between the scene image $I_{RGB}^t$ and each target object $o_{t_j}$ is computed as the cosine similarity between their respective features. This is achieved using the BLIP-2 VLM [43]. Specifically, the image-text matching model of BLIP-2 is utilized to compute the contrastive score, which serves as their cosine similarity. We selected BLIP-2 as it provides semantically rich embeddings for multi-modalities, necessary for capturing the relationship between objects and scenes.

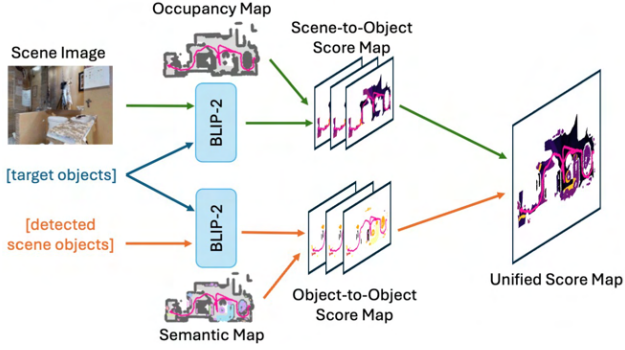

**Fig. 3.** Overview of the Unified Score Map generation process.

We follow [23] by generating a cone-shaped confidence mask $C^t$ at each time step to represent the camera's field of view (FOV). The confidence of each pixel is maximal at the optical axis with a value of 1 and decreases away from the optical axis based on $\cos^2(\theta/(\theta_{FOV}/2) * \pi/2)$. Pixels representing obstacles, identified from $I_D^t$, are assigned a value of 0 in $C^t$. Each channel in the scene-level score map $V_{S \to O_{tgt}}^t$, namely $V_{S \to o_{t_j}}^t \in \mathbb{R}^{1 \times H \times W}$, corresponds to the score map for object $o_{t_j}$, and can be obtained through two steps. First, by scaling $C^t$ with $S(I_{RGB}^t, o_{t_j})$ to obtain $\widetilde{V}_{S \to o_{t_j}}^t$:

$$\widetilde{V}_{S \to o_{t_j}}^t = C^t \cdot S(I_{RGB}^t, o_{t_j}), \forall o_{t_j} \in \mathcal{O}_{tgt}. \quad (2)$$

Then, by updating based on a weighted average of the current and previous values [23]:

$$V_{S \to o_{t_j}}^t = \frac{C^t \odot \widetilde{V}_{S \to o_{t_j}}^t + C^{t-1} \odot V_{S \to o_{t_j}}^{t-1}}{C^t + C^{t-1}}, \forall o_{t_j} \in \mathcal{O}_{tgt}, \quad (3)$$

where $\odot$ is the Hadamard product. Similarly, the confidence map $C^t$ is updated as follows [23]:

$$C^t = \frac{(C^t)^2 + (C^{t-1})^2}{C^t + C^{t-1}}. \quad (4)$$

*2) Object to Object Score Mapping*

The *OtO Score Mapping* module generates a score map representing fine-grained, object-level correlations between target objects and scene objects, Fig. 3. Each element in the score map represents the cooccurrence score of a specific location in the scene. Specifically, a higher score represents the presence of scene objects that commonly appear with the target objects. It takes as inputs $I_{RGB}^t$ and $M_S^t$, and outputs a multi-channel scene object to target object score map $V_{O_{sne} \to O_{tgt}}^t \in \mathbb{R}^{K \times H \times W}$ of the same spatial dimension as $M_o^t$.

We compute a cooccurrence matrix $W \in \mathbb{R}^{L \times K}$ where $w_{ij} = S(e_{s_i}, e_{t_j}) \in W$ represents the cosine similarity between the embeddings of the scene object $o_{s_i}$ and target object $o_{t_j}$. These embeddings are encoded by the Q-Former of BLIP-2. For each target object $o_{t_j}$, the corresponding channel of the score map $V_{O_{sne} \to o_{t_j}}^t$ is computed by weighting each channel $i$ of the semantic map, $M_{S,o_{s_i}}^t$, representing the presence of a scene object $o_{s_i}$, with the cosine similarity $S(e_{s_i}, e_{t_j})$. This scales the contribution of each scene object by how semantically correlated it is to the target object. The OtO score map for each target object $o_{t_j}$ at each time step $t$ is then given by:

$$V_{O_{sne} \to o_{t_j}}^t = \sum_{\mathcal{O}_{sne}} M_{S,o_{s_i}}^t S(e_{s_i}, e_{t_j}), \forall o_{t_j} \in \mathcal{O}_{tgt}. \quad (5)$$

*3) Score Fusion*

The *Score Fusion* module introduces a score fusion technique that combines both scene- and object-level correlations into a unified score map to guide the robot towards regions of high target object likelihood. Specifically, it combines the multi-channel StO score map $V_{S \to o_{t_j}}^t$ and OtO score map $V_{O_{sne} \to o_{t_j}}^t$. Before fusion, each per-target channel is independently min-max normalized to prevent any single object from dominating the final score map. The unified score map $V_{S, O_{sne} \to O_{tgt}}^t \in \mathbb{R}^{H \times W}$ is obtained by element-wise addition of the normalized score maps $V_{S \to o_{t_j}}^t$ and $V_{O_{sne} \to o_{t_j}}^t$, and then summing over the channels to obtain a combined score:

$$V_{S, O_{sne} \to O_{tgt}}^t = \sum_{\mathcal{O}_{tgt}} V_{S \to o_{t_j}}^t + V_{O_{sne} \to o_{t_j}}^t. \quad (6)$$

Because our experiments assume equal priority for all targets, we fuse by simple summation. Therefore, locations on the unified score map, Fig. 3, that are semantically relevant to multiple targets objects, and/or locations with scene objects that are semantically relevant to multiple target objects, will accumulate higher scores. The unified score map $V_{S, O_{sne} \to O_{tgt}}^t$ is then passed into the *Exploration Planner* module.

*D. Exploration Planner*

The *Exploration Planner* selects the next frontier $g$ or target object waypoint $w_{t_i}$ to navigate towards. It comprises two sub-modules: 1) *Frontier Selection*, and 2) *Navigation Controller*. If no target object is detected by the *Object Detection* sub-module, the *Frontier Selection* sub-module determines the next frontier $g$ to explore. If a target object $o_{t_i}$ is detected, then the *Navigation Controller* directly receives the target waypoint $w_{t_i}$ and navigates towards it. When the distance between the robot and the detected target object is within a threshold $\epsilon$, the target object is found, e.g., $o_{t_j}$, the object is removed from the search list $\mathcal{O}_{tgt}$, e.g., $\mathcal{O}_{tgt} = \mathcal{O}_{tgt} \setminus \{o_{t_j}\}$. When the search list is empty, the robot triggers "stop" action.

*1) Frontier Selection*

The *Frontier Selection* sub-module selects the next frontier $g(x, y)$ for the robot to explore using the occupancy map $M_o^t$ and unified score map $V_{S, O_{sne} \to O_{tgt}}^t$. Frontier points are defined as the midpoint at each boundary separating the explored and unexplored areas [44]. The frontier $g$ with the highest unified score is passed into the *Navigation Controller*.

*2) Navigation Controller*

The *Navigation Controller* sub-module generates robot control actions $u$ using either the target object waypoint $\mathbf{w}_{t_i}$ from the *Object Detection* sub-module, or the frontier $g$ from the *Frontier Selection* sub-module. We use a point goal navigation policy Variable Experience Rollout (VER) [45] pretrained in [23]. Robot actions include "move forward", "turn left", "turn right", and "stop".

## V. EXPERIMENTS

We conducted four sets of experiments to evaluate the overall performance of Finder on the MOS task: 1) a comparison study against state-of-the-art (SOTA) methods in simulated indoor environments, 2) an ablation study to investigate the impact of StO and OtO score map on multi-object search efficiency, 3) a scalability study to evaluate the impact of increasing the number of search targets on exploration time, and 4) a sim-to-real study in an indoor multi-area building environment to evaluate the generalizability of Finder to real-world environments.

### A. Simulation Comparison Study

We compared Finder against SOTA methods using the Habitat simulator [46] on two datasets consisting of high-resolution 3D scans from real-world buildings: HM3D [47], and MP3D [48]. For both HM3D and MP3D datasets, we ran 1000 episodes per method. At the beginning of each episode, the robot was spawned at a random location inside the environment and given a list of 3 target objects. An episode terminated if the robot triggered "stop" or the total number of time steps exceeded 500.

*1) Procedure:* We used two performance metrics for these experiments: Success rate (SR), to measure the percentage of successful episodes where the robot found all target objects, and multi-object success weighted by normalized inverse path length (MSPL), based on SPL [49], which measures multi-object search efficiency. MSPL is calculated by:

$$\text{MSPL} = \frac{1}{N} \sum_{i=1}^{N} S_i \frac{\ell_i}{\max(p_i, \ell_i)}, \quad (7)$$

where $N$ denotes the number of episodes, $S_i$ is a binary indicator of success of episode $i$, $\ell_i$ denotes the optimal shortest path length from the start location to all target objects, and $p_i$ denotes the actual robot path length.

*2) Comparison Methods:* We compared against the following three sets of methods: DRL, VLM, and Lower and Upper Bounds.

*a) DRL method*

In terms of DRL methods, we evaluated our approach against a seminal work in MOS.

**Multi-Object Navigation (MultiON) [7]:** MultiON uses RGB-D images, a goal vector, and a metric map as inputs. The model uses a ConvNet to process visual inputs and a GRU [50] to maintain memory of the robot's state for action generation. MultiON remains the only open-source DRL method that addresses MOS end-to-end, making it our sole DRL baseline.

*b) VLM methods*

In terms of VLM methods, we compared Finder against four methods that use visual or language embeddings to guide the search process. The following methods were selected as 1) they used the same sensor input as ours, 2) are open sourced, and 3) widely recognized in the research community. These methods were originally designed for SOS; however, we adapted them for MOS by searching for the target object from $\mathcal{O}_{\text{tgt}}$ sequentially. To ensure fair comparison, we used a minimal adaptation strategy by resetting the target prompt after each found object so that each baseline's original policy and planning logic remained unchanged.

**CLIP on Wheels (CoW)** [20]: CoW constructs a metric map from RGB-D images for frontier exploration and uses CLIP to localize the target object.

**Leveraging Large Language Models for Visual Target Navigation (L3MVN Zero-Shot)** [27]: L3MVN (Zero-Shot) builds a semantic map from RGB-D inputs and uses LLMs to score frontiers from the semantic map for waypoint selection.

**L3MVN (Feed-Forward)** [27]: L3MVN (Feed-Forward) uses a feed-forward network to predict frontiers from the semantic map based on LLM embeddings.

**Vision-Language Frontier Maps (VLFM)** [23]: VLFM generates a value map based on the cosine similarity between the scene image and the target object for frontier selection.

*c) Lower and Upper Bound methods*

We compared against a lower and upper bound approach to evaluate Finder's performance in relation to baseline and optimal strategies.

**Random Walk (lower bound)**: The robot randomly selects a navigation action at each timestep. It serves as the lower bound approach.

**Oracle (upper bound)**: Oracle plans an optimal shortest path to all the target objects given access to the ground-truth of the object locations and the map. It serves as an upper bound approach.

*3) Results:*

The quantitative results of comparison study are presented in Table I. Finder outperformed Random Walk, MultiON, CoW, L3MVN, VLFM in terms of SR and MSPL on both HM3D and MP3D datasets. Finder achieved higher SR and MSPL than CoW because CoW only used VLMs to localize the target object. Specifically, CoW did not incorporate reasoning about frontier selection based on the semantic relationship between the scene and the target, leading to less efficient object searches. Similarly, Finder outperformed L3MVN by integrating visual observations and generating a unified score map, while L3MVN relied solely on language semantic priors. In comparison to VLFM, Finder's higher performance is attributed to its consideration of both scene-level and object-level correlations between the environment and the target object. On the MP3D dataset, Finder also outperformed MultiON, which used predefined cylinders as target objects, disregarding semantic relationships with the robot's environment. Finder achieved lower SR and MSPL in the MP3D dataset compared to the HM3D dataset because part of the scenes in the MP3D dataset are larger indoor environments. They require longer travel time for all target objects to be found, resulting in lower SR and MSPL given the same amount of maximum timesteps in each episode.

TABLE I
COMPARISON BETWEEN FINDER AND SOTA METHODS

| Methods | HM3D SR↑ | HM3D MSPL↑ | MP3D SR↑ | MP3D MSPL↑ |
|---|---|---|---|---|
| Random Walk | 0.5% | 0.0043 | 0.0% | 0.0 |
| MultiON | - | - | 23.9% | 0.159 |
| CoW | 14.2% | 0.113 | 1.9% | 0.059 |
| L3MVN (Zero-Shot) | 27.2% | 0.187 | 6.6% | 0.043 |
| L3MVN (Feed-Forward) | 28.1% | 0.188 | 7.3% | 0.051 |
| VLFM | 32.4% | 0.155 | 12.6% | 0.104 |
| Oracle | 100.0% | 1.0 | 100.0% | 1.0 |
| **Finder (ours)** | **63.4%** | **0.389** | **55.4%** | **0.344** |

TABLE II
ABLATION STUDY

| | SR↑ | MSPL↑ |
|---|---|---|
| Finder w/o StO | 61.5% | 0.364 |
| Finder w/o OtO | 58.3% | 0.337 |
| **Finder (ours)** | **63.4%** | **0.389** |

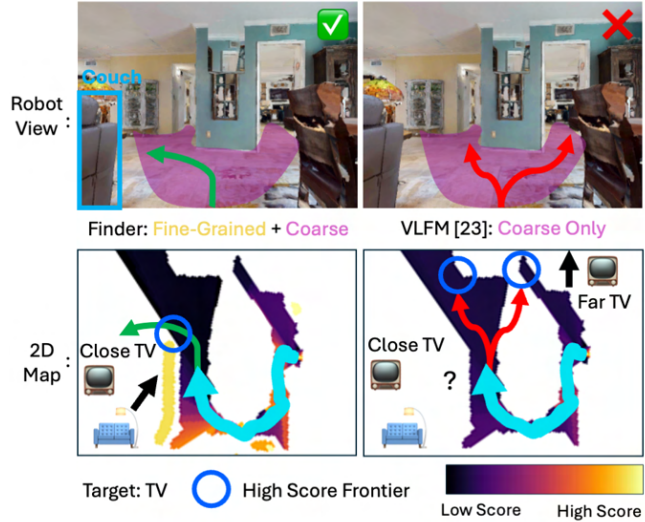

**Fig. 4.** (left) When searching for the target object TV, Finder generates both fine-grained and coarse score maps to recognize that TVs are semantically found close to couches. This allows the robot to take the shortest route to a closer TV. (right) VLFM [23] generates only coarse scores which considers a sparser set of scores. This leads the robot to a longer route to find the farther TV.

The statistical significance of Finder's performance was established through a two-step analysis. 1) A Friedman test to show significant differences among SR and MSPL of all methods ($p < 0.001$), followed by 2) Post-hoc Wilcoxon signed-rank tests with Bonferroni correction comparing Finder against each baseline method (except Oracle). The Wilcoxon tests showed that Finder significantly outperformed all baseline methods across both HM3D and MP3D datasets ($p < 1 \times 10^{-29}$).

A qualitative test shows how Finder can achieve optimal outcomes using a combination of fine-grained and coarse scores when compared against VLFM, Fig. 4. Two trials were conducted under identical initial conditions: Finder utilized both StO and OtO score maps, while VLFM used only the StO score map. Finder's combined 2D score map allocated high scores toward the detected couch after comparing the target (TV) with the scene object (couch) in addition to subtle scene scores. In future timesteps, this allowed the robot to identify the optimal search area for the TV to be near a couch, which led to initiating a left turn. In comparison, VLFM, relying solely on coarse scene-context scores, assigned frontiers to less relevant areas of the scene based on its point-of-view. Future timesteps confirmed that VLFM took a longer route to locate the target object, which is indicated as "Far TV".

*B. Simulation Ablation Study*

We conducted an ablation study to investigate the impact of the different score maps used in Finder on multi-object search performance. Namely, we considered the following two variants: **1) Finder w/o Scene-to-Object score map**: This variant does not include StO score map for frontier selection; and **2) Finder w/o Object-to-Object score map**: This variant does not include OtO score map for frontier selection. We conducted 1000 episodes per method using the HM3D dataset, following the procedure in Section V. A.

*1) Results:* The ablation study results are presented in Table II. The full Finder system achieved a SR of 63.4% and an average MSPL of 0.389. In contrast, removing the scene-level object correlations (Finder w/o StO) caused a decrease in performance, with an SR of 61.5%, and an MSPL of 0.364. Without the scene-level correlations, the robot disregarded areas that were semantically correlated to the target objects. For example, the robot might skip exploring a kitchen-like area when searching for a toaster. Similarly, removing object-to-object correlations (Finder w/o OtO) further reduced the SR to 58.3% and the MSPL to 0.337. Without these correlations, the robot could not exploit the cooccurrence of objects that typically appear together. For instance, when searching for a TV, the robot might miss areas with a remote control or couch, which are often found near TVs. Thus, the absence of these score maps resulted in a degraded understanding of the semantic relationships between scene objects, target objects and the environment, leading to reduced search performance.

*C. Simulation Scalability Study*

*1) Procedure:* We evaluated the performance of Finder in terms of exploration time for increasing number of target objects. The objective is to investigate Finder's efficiency as task complexity grows. We conducted 100 successful episodes for each experimental condition, varying the number of target objects from 1 to 8, using the HM3D dataset.

*2) Results:* The results of scalability study are presented in Fig. 5. (a). Overall, the average exploration time increased as the number of target objects increased. Exploration time increased from 67 steps to over 200 steps as the number of target objects exceeded one, indicating that the search task becomes significantly more complex when transitioning from SOS to MOS. However, the exploration time gradually converged to around 300 steps, with only marginal increases as the number of objects increased. This convergence suggests that Finder effectively explores a substantial portion of the environment within this time, enabling it to find all target objects efficiently. These results demonstrate Finder's capability to scale in MOS tasks.

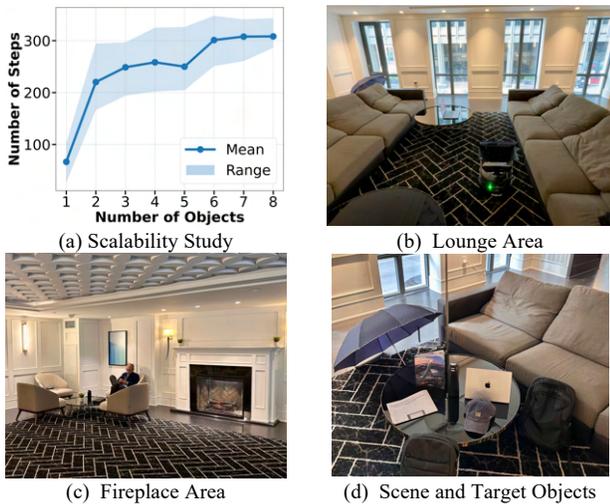

**Fig. 5.** (a) The scalability study results. (b)-(d) multi-room indoor environment with varying areas, as well as scene and target objects.

*D. Sim-to-Real Study*

We conducted real-world experiments in an object-dense multi-area indoor building environment with a total area of 121.5 $m^2$, Fig. 5. (b)-(c). Namely, it consists of lounge area, study area, and a fireplace area. A TurtleBot2 was deployed with a Kinect camera for obtaining RGB-D image observations. Robot Operating System (ROS) Noetic was used on the TurtleBot2's onboard computer. We ran BLIP-2, Grounding DINO, and MobileSAM on a separate desktop with an NVIDIA RTX 4090 GPU over internet connection. We used a set of target objects including garbage bin, fireplace, laptop, shoes, backpack, lamp, and umbrella, Fig. 5. (d). We sampled 3, 4, 5 objects from the set of target object lists for each trial to evaluate: 1) the generalizability of Finder in real-world environments, and 2) its ability to find increasing number of objects. For the navigation controller, we used A* as the global planner and Time Elastic Band Planner [51] as the local planner to generate robot velocities to navigate to the selected waypoint. Object detector module ran at 2 to 3 Hz over a local network with around 400 ms latency per frame. A video of Finder addressing the MOS task in both simulated and real-world environments is provided on https://find-all-my-things.github.io/.

## VI. CONCLUSION

In this paper, we introduced Finder, a novel VLM-based approach to address the multi-object search problem across various environments. The proposed method integrates multi-channel Scene-to-Object and Object-to-Object score maps generated from VLMs for effective waypoint selecting during object search. These score maps enable simultaneous tracking and reasoning about multiple objects, while leveraging both scene-level and object-level semantic correlations. Extensive experiments were conducted in simulated and real-world environments against SOTA methods. The results demonstrated that Finder outperformed existing multi-object search methods. Ablation study further confirmed the effectiveness of our multi-channel score maps and fusion technique, while scalability study demonstrated Finder's performance with increasing number of target objects. The limitations of Finder include reliance on accurate object detection, static indoor environments, and high-performance compute. Future work includes extending Finder to handle dynamic objects and interactive search scenarios where objects may be hidden, moved or stored. We also aim to compress object detection models for low-latency inference on edge devices.